\documentclass[11pt]{article}
\usepackage[hyperref]{ccl2023-en}
\usepackage{times}
\usepackage{url}
\usepackage{latexsym}
\usepackage{fancyhdr}
\usepackage[namelimits]{amsmath} 
\usepackage{amssymb}             
\usepackage{amsfonts}            
\usepackage{mathrsfs}            
\usepackage{multirow}
\usepackage[normalem]{ulem}
\useunder{\uline}{\ul}{}
\usepackage{microtype}
\usepackage{booktabs}
\usepackage{algorithm2e}
\usepackage{CJK}
\usepackage{inconsolata}
\usepackage{ulem}
\usepackage{makecell}
\usepackage{multirow}
\usepackage{amsmath}
\usepackage{amssymb}
\usepackage{enumitem}
\usepackage{color}
\usepackage{comment}
\usepackage{setspace}
\usepackage{geometry}
\usepackage{array}
\usepackage{graphicx}
\usepackage {inconsolata}
\usepackage{hyperref}
 
\setenumerate[1]{itemsep=0pt,partopsep=0pt,parsep=\parskip,topsep=0pt}
\setitemize[1]{itemsep=0pt,partopsep=0pt,parsep=\parskip,topsep=0pt}
\title{A Systematic Evaluation of Large Language Models for Natural Language Generation Tasks}

\author{Xuanfan Ni, Piji Li\thanks{$^*$Corresponding author.} \\ College of Computer Science and Technology,\\Nanjing University of Aeronautics and Astronautics\\
MIIT Key Laboratory of Pattern Analysis and Machine Intelligence
\\{\tt \{xuanfanni, pjli\}@nuaa.edu.cn}} 

\date{}

\begin{document}
\maketitle
\begin{abstract}
Recent efforts have evaluated large language models (LLMs) in areas such as commonsense reasoning, mathematical reasoning, and code generation. However, to the best of our knowledge, no work has specifically investigated the performance of LLMs in natural language generation (NLG) tasks, a pivotal criterion for determining model excellence. Thus, this paper conducts a comprehensive evaluation of well-known and high-performing LLMs, namely ChatGPT, ChatGLM, T5-based models, LLaMA-based models, and Pythia-based models, in the context of NLG tasks. We select English and Chinese datasets encompassing Dialogue Generation and Text Summarization. Moreover, we propose a common evaluation setting that incorporates input templates and post-processing strategies. Our study reports both automatic results, accompanied by a detailed analysis.
\end{abstract}

\section{Introduction}
\label{intro}

%
%
\cclfootnote{
    %
    %
    \hspace{-0.65cm}  
    \textcopyright 2023 China National Conference on Computational Linguistics

    \noindent Published under Creative Commons Attribution 4.0 International License
}

Recent studies have emphasized the importance of scaling large language models (LLMs), referring to both the dimensions of the model size themselves and the amount of data used, resulting in enhanced capability of the models for tasks downstream \cite{DBLP:journals/corr/abs-2210-11416}. Numerous investigations have been conducted to explore the limits of performance by training increasingly larger pre-trained language models, such as GPT-3 175B \cite{DBLP:journals/corr/abs-2005-14165} and PaLM 540B \cite{DBLP:journals/corr/abs-2204-02311}. Although scaling primarily involves increasing the model size while maintaining similar architectures and pre-training tasks, these large-sized PLMs exhibit distinct behaviors from their smaller counterparts and demonstrate surprising \textbf{emergent abilities} in solving complex tasks \cite{DBLP:conf/iclr/ZhangBHRV17,DBLP:conf/iclr/FrankleC19,DBLP:journals/cacm/ZhangBHRV21}. An example of this is the contrasting performance of GPT-3 and GPT-2 when it comes to solving few-shot tasks. GPT-3 demonstrates effective problem-solving abilities by utilizing in-context learning, whereas GPT-2 faces difficulties in this aspect. As a result, these large-scale language models (LLMs) has become a huge research topic in current NLP area. In existing literature, remarkable LLMs such as ChatGPT\footnote{https://chat.openai.com/}, ChatGLM\footnote{https://chatglm.cn/}, have been widely adopted as powerful AI assistants, benefiting from their exceptional generation capabilities.

We hypothesis that a language model's performance in executing natural language generation (NLG) tasks is a crucial factor in determining its excellence \cite{DBLP:journals/csur/DongLGCLSY23}. NLG tasks involve LLMs that are capable of accepting diverse types of input, such as texts and tables, and generating coherent and appropriate output text. We intuitively think that generate fluent, coherent, and consistent texts is the foundation of a language model, so as to large language models \cite{DBLP:journals/jmlr/RaffelSRLNMZLL20}. When some research institutions release their large language models, they tend to evaluate these models first. Community workers are also interested in testing well-known large language models. However, most of these evaluations focus on checking LLMs' ability of commonsense reasoning \cite{DBLP:journals/cacm/DavisM15,DBLP:conf/nips/Wei0SBIXCLZ22}, mathematical reasoning \cite{DBLP:conf/iclr/SaxtonGHK19,DBLP:conf/nips/Wei0SBIXCLZ22}, code completion \cite{DBLP:journals/csur/AllamanisBDS18}, etc., but ignore the basic NLG tasks, such as dialogue generation \cite{DBLP:journals/sigkdd/ChenLYT17}, text summarization \cite{DBLP:journals/csur/DongLGCLSY23}, and story generation \cite{DBLP:journals/csur/AlhussainA21}. Besides, Some researchers pointed out that the performance of a large model is determined not only by its size and architecture, but more by the quality and quantity of training data. Based on this point of view, researchers open source and propose that some smaller-scale models trained on more and higher-quality data sets can achieve the same performance as models with more parameters than them. For example, LLaMA-13B \cite{DBLP:journals/corr/abs-2302-13971} outperforms GPT-3 on most benchmarks, despite being 10 times smaller. This notable discovery makes us curious about the performance of models with different architecture, data size, and mode size, trying to figure out which factor is more important. Therefore, we aim to address this gap by conducting a comparative analysis of LLM performance on NLG tasks, considering different architectures and scales throughout the evaluation process.

In this paper, we present a systematic evaluation of existing LLMs for NLG tasks. The main objective is to enhance our understanding of instruction and prompt design by conducting a comparative analysis of these models. Initially, we provide an overview of classic NLG tasks, including their definitions and associated English and Chinese datasets. Subsequently, we devise a model input template that includes instructions for each dataset. Following that, we introduce various LLMs, considering factors such as model size and architecture. Finally, we present the results of both automatic and manual evaluation of LLMs on NLG datasets, and discuss the strengths and weaknesses of their performance across different models.


\section{Natural Language Generation}
In this section, we will introduce the definition of NLG, and its sub-tasks with some corresponding datasets that we will use to evaluate LLMs.
\subsection{Definition}
Natural Language Generation is the process of producing a natural language text in order to meet specified communicative goals. The texts that are generated may range from a single phrase given in answer to a question, through multi-sentence remarks and questions within a dialog, to full-page explanations. In our evaluation, we mainly focus on text-to-text styles. 
In general, the task of NLG targets at finding an optimal sequence $y_{<T+1}=(y_1,y_2,\dots,y_T)$ that satisfies:
\begin{equation}
    y_{<T+1}=\underset{y_{<T+1} \in \mathcal{Y}}{\arg \max } \log P_{\theta}\left(y_{<T+1} \mid x\right)=\underset{y_{<T+1} \in \mathcal{Y}}{\arg \max } \sum_{t=1}^{T} \log P_{\theta}\left(y_{t} \mid y_{<t}, x\right)
\end{equation}
where $T$ represents the number of tokens of the generated sequence, $\mathcal{Y}$ represents a set containing all possible sequences, and $P_{\theta}\left(y_{t} \mid y_{<t}, x\right)$ is the conditional probability of the next token $y_t$ based on its previous tokens $y_{<t}=(y_1,y_2,\dots,y_{t-1})$ and the source sequence $x$ with model parameters $\theta$.

Next, we will introduce some classic and widely-researched sub-tasks of NLG, with several corresponding datasets.

\subsection{Dialogue Generation}
Dialogue generation refers to the process of automatically generating coherent and contextually appropriate responses in a conversational setting \cite{DBLP:journals/sigkdd/ChenLYT17,DBLP:journals/inffus/MaNXC20,DBLP:journals/csur/DongLGCLSY23}. The ultimate goal of dialogue generation task is to create responses that are relevant, informative, and engaging to the user.We utilize two English dialogue datasets characterized by clear emotional flow and topic constraints, as well as one English dataset that incorporates speakers' personalities. Furthermore, we employ a Chinese open-domain dialogue dataset for evaluation purposes.
\begin{itemize}
    \item \textbf{DailyDialog} \cite{DBLP:conf/ijcnlp/LiSSLCN17} is a comprehensive, human-authored, and relatively noise-free English dataset that captures everyday communication styles and encompasses various topics related to our daily lives.
    \item \textbf{PersonaChat} \cite{DBLP:conf/acl/KielaWZDUS18} is a persona-grounded dialogue dataset which contains 10k English multi-turn dialogues conditioned on personas, and each persona is described with at least 5 profile sentences.
    \item \textbf{EmpatheticDialogue} \cite{DBLP:conf/acl/RashkinSLB19} is a large-scale multi-turn dialogue English dataset that contains 25k empathetic conversations between a speaker and a listener.
    \item \textbf{LCCC} \cite{DBLP:conf/nlpcc/WangKZHJZH20} is a large-scale cleaned Chinese conversation dataset.
\end{itemize}

\subsection{Text Summarization}
Text summarization is the process of condensing a piece of text, such as an article, document, or news story, into a shorter version while preserving its key information and main ideas \cite{DBLP:journals/eswa/El-KassasSRM21,DBLP:journals/csur/DongLGCLSY23}. Text summarization can be performed through two main approaches: \textit{Extractive Summarization}  and \textit{Abstractive Summarization}. In our evaluation, we utilize multiple abstractive summarization datasets, specifically choosing two renowned datasets for the English and Chinese languages.
\begin{itemize}
\item \textbf{CNN/DailyMail} \cite{DBLP:conf/conll/NallapatiZSGX16} is a large scale English summarization dataset which contains 93k and 220k articles collected from the CNN and Daily Mail websites, respectively, where each article has its matching abstractive summary.
\item \textbf{XSum} \cite{DBLP:conf/emnlp/NarayanCL18} is an extreme English summarization dataset containing BBC articles and corresponding single sentence summaries. In this dataset, 226,711 Wayback archived BBC articles are collected, which range from 2010 to 2017 and cover a wide variety of domains.
\item \textbf{THUCNews} \cite{DBLP:conf/emnlp/LiS07} is a Chinese summarization dataset, which comes from filtering the historical data of the Sina News RSS subscription channel from 2005 to 2011, including 740,000 news documents.
\item \textbf{LCSTS} \cite{GPT2-NewsTitle} is a large corpus of Chinese short text summarization dataset constructed from the Chinese micro-blogging website \textit{Sina Weibo}. This corpus consists of over 2 million real Chinese short texts with short summaries given by the author of each text.  
\end{itemize}



\section{Experimental Settings}
\begin{figure*}[!t]
   \centering
   \includegraphics[width=\linewidth]{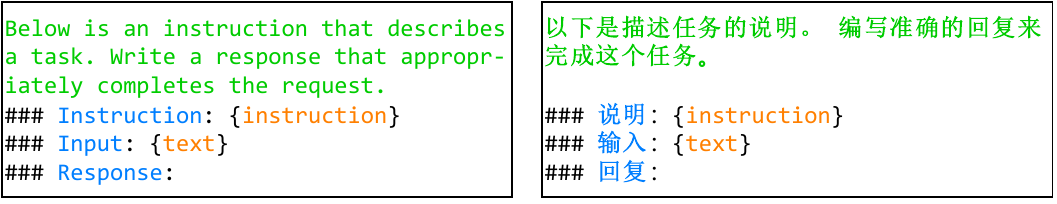}
        \caption{Input templates for English (left) and Chinese (right) datasets. \textbf{instruction} and \textbf{text} will be replaced with content corresponding different datasets.}
    \label{template}
\end{figure*}

\subsection{Overview for LLMs}
Typically, large language models (LLMs) refer to Transformer-based models containing tens or hundreds of billions of parameters and trained on extensive corpora of texts \cite{DBLP:journals/corr/abs-2303-18223}. These LLMs demonstrate significant capabilities in understanding natural language and solving complex tasks. Furthermore, they have showcased their ability to perform new tasks based on textual instructions or with just a few examples \cite{DBLP:journals/corr/abs-2210-11416}. The emergence of these few-shot properties is a result of scaling models to a sufficient size, leading to a line of research that focuses on further scaling these models \cite{DBLP:journals/corr/abs-2112-11446}.

Previous LLMs, such as T5 \cite{DBLP:journals/jmlr/RaffelSRLNMZLL20}, GPT-3 \cite{DBLP:journals/corr/abs-2005-14165}, OPT \cite{DBLP:journals/corr/abs-2205-01068}, and PaLM \cite{DBLP:journals/corr/abs-2204-02311}, primarily emphasized scaling model size rather than considering the quality and quantity of data. However, recent studies have demonstrated that, given a fixed compute budget, the best performance is achieved by smaller models trained on larger datasets \cite{DBLP:journals/corr/abs-2203-15556}. Additionally, most of these models are not open-source and can only be accessed through APIs for inference, which poses inconveniences for model evaluation and usage. In order to address this issue, numerous researchers have proposed excellent open-source architectures and trained models, including GLM-130B \cite{DBLP:journals/corr/abs-2210-02414}, ChatGLM \cite{DBLP:conf/acl/DuQLDQY022}, LLaMA \cite{DBLP:journals/corr/abs-2302-13971}, and Pythia \cite{DBLP:journals/corr/abs-2304-01373}. Furthermore, advancements in fine-tuning techniques have contributed to the success of deploying these models with limited resources, such as Lora \cite{DBLP:conf/iclr/HuSWALWWC22} and P-Tuning \cite{DBLP:conf/acl/LiL20}. Therefore, this paper aims to conduct systematic evaluations of these models and their fine-tuned versions, categorized into four groups: \textbf{ChatGPT, ChatGLM, T5-based models, LLaMA-based models, and Pythia-based models}.

\begin{table*}[!t]
\centering
\resizebox{\linewidth}{!}{
\begin{tabular}{m{2cm}|m{10.5cm}|m{3cm}}
\toprule[1.3pt]
\multicolumn{1}{c|}{\textbf{Dataset}} & \multicolumn{1}{c|}{\textbf{Instruction}}  & \multicolumn{1}{c}{\textbf{Text}}                       \\ \cmidrule{0-2}
Empathetic Dialogues        & This is an open-domain \textcolor{red}{\textit{empathetic}} dialogue completion task.The input is the Dialogue. You act as System in the dialogue. You need to fully \textcolor{blue}{\textit{understand the situation and combine the speaker's emotion}} to complete the dialogue with natural content and a way closer to human speech. There is no need for any additional notes or clarifications, you just give the response in English. & Dialogue Context\\ \cmidrule{0-2}
DailyDialog                 & This is an open-domain \textcolor{red}{\textit{topic-aware}} dialogue completion task. The input is the Dialogue. You act as System in the dialogue. You need to fully \textcolor{blue}{\textit{understand the topic}} and complete the dialogue with natural content and a way closer to human speech. There is no need for any additional notes or clarifications, you just give the response in English                                     & Dialogue Context \\ \cmidrule{0-2}
PersonaChat                 & This is an open-domain \textcolor{red}{\textit{personality-aware}} dialogue completion task. The input is the Dialogue. You act as System in the dialogue. You need to fully \textcolor{blue}{\textit{understand the personality}} and complete the dialogue with natural content and a way closer to human speech. There is no need for any additional notes or clarifications, you just give the response in English.                        & Dialogue Context \\ \cmidrule{0-2}
LCCC                        & \begin{CJK*}{UTF8}{gbsn}这是一个开放域的中文对话补全任务。输入是待完成的对话内容。 你在对话中扮演系统。你需要完全理解说话者的话语，并用自然的内容和更接近于人类说话的方式完成对话，而不是用语言模型或者AI的身份。不需要任何额外的注释或者说明，你只需用中文给出回复。\end{CJK*} & Dialogue Context \\ \bottomrule[1.3pt]
\end{tabular}
}
\caption{\label{input}\textit{Instruction} and \textit{Text} for each dataset.}
\end{table*}

\subsection{ChatGPT}
ChatGPT\footnote{https://chat.openai.com/} is a large language model based on OpenAI's GPT-3.5 architecture \cite{DBLP:journals/corr/abs-2005-14165}. It is designed specifically for generating conversations and answering user queries. ChatGPT employs large-scale pretraining and fine-tuning methodologies, utilizing vast amounts of textual data to learn statistical patterns and semantic knowledge of language, and perform well in zero-shot and few-shot settings, and can understand the input instructions.

\subsection{ChatGLM}
ChatGLM\footnote{https://chatglm.cn/} is a freely available dialogue language model that operates in both Chinese and English languages. It follows the GLM architecture and boasts an impressive parameter count of 6.2 billion. ChatGLM-6B incorporates similar technology as ChatGPT, with a specific focus on Chinese question answering and dialogue. The model undergoes extensive training on a dataset containing approximately 1 trillion tokens in Chinese and English. The training process includes supervised fine-tuning, feedback bootstrap, and reinforcement learning with human feedback. Despite having only 6.2 billion parameters, the model demonstrates the ability to generate responses that align with human preferences.

\begin{table}[!t]
\centering
\resizebox{\linewidth}{!}{
\begin{tabular}{l|l|l|l|lll|l|l|ll|l}
\toprule[1.3pt]
\textbf{Model}          & \textbf{Scale} & \textbf{Arch} & \textbf{PPL\textdownarrow}    & \textbf{B-1} & \textbf{B-2} & \textbf{B-4} & \textbf{MT} & \textbf{R-L }& \textbf{D-1} & \textbf{D-2} & \textbf{PPR\textdownarrow}     \\ \cmidrule{0-11}
\textsc{EP-PG} & -- & -- & -- & 16.74 & 6.94 & 2.39 & -- & -- & 2.19 & 8.25 & -- \\
MoEL & 23.1M & DO & 33.58 & -- & -- & 2.90 & -- & -- & 1.06 & 4.29 & -- \\ \cmidrule{0-11}
ChatGPT        & 175B  & DO   & 10.52  & 7.35   & 2.40   & 0.52   & \textbf{9.26}   & \textbf{8.75}   & 4.71   & 27.75  & \textbf{0.00\% }    \\ \cmidrule{0-11}
ChatGLM        & 6B    & DO   & 11.73  & 6.05   & 1.82   & 0.27   & 8.58   & 7.71   & 3.57   & 22.82  & 12.61\% \\ \cmidrule{0-11}
Flan-T5-XXL    & 13B   & ED   & 19.97  & 5.62   & 2.40   & \underline{0.61}   & 5.38   & 7.41   & 5.66   & 24.97  & \textbf{0.00\% }    \\
FastChat-T5    & 3B    & ED   & \underline{9.25}   & 7.33   & 2.35   & 0.45   & 8.50   & 8.62   & 3.55   & 20.81  & 0.12\%  \\ \cmidrule{0-11}
Open-LLaMA     & 7B    & DO   & 15.90  & \textbf{8.50}   & \textbf{2.97}   & \textbf{0.63}   & 6.43   & \underline{8.74}   & 3.93   & 17.91  & 40.05\% \\ 
Vicuna         & 13B   & DO   & 14.31  & 6.18   & 1.93   & 0.35   & \underline{8.91}   & 7.81   & 4.09   & 25.84  & 38.86\% \\ 
Alpaca-Lora    & 7B    & DO   & 16.10  & 7.95   & 2.52   & 0.40   & 7.34   & 6.69   & \textbf{7.59}   & \underline{39.58}  & 0.24\%  \\ 
Chinese-Alpaca & 13B   & DO   & 12.05  & 6.51   & 1.86   & 0.35   & 7.53   & 6.64   & 5.32   & 29.14  & 0.20\%  \\ 
GPT4ALL        & 13B   & DO   & 11.14  & 5.20   & 1.47   & 0.24   & 8.75   & 6.78   & 3.94   & 25.60  & 1.81\%  \\ \cmidrule{0-11}
Dolly          & 12B   & DO   & 131.75 & \underline{8.29}   & \underline{2.64}   & 0.46   & 6.91   & 7.96   & \underline{7.46}   & \textbf{42.69}  & 58.61\% \\ 
Oasst-Pythia   & 12B   & DO   & \textbf{8.71}   & 5.48   & 1.53   & 0.26   & 8.79   & 6.92   & 3.38   & 21.18  & \underline{0.04\%} \\ \bottomrule[1.3pt]
\end{tabular}
}
\caption{Automatic evaluation results of LLMs on EmpatheticDialogues. \textbf{Scale} stands for the model size.\textbf{ED} and \textbf{DO} respectively stand for \textit{encoder-decoder} and \textit{decoder-only}. \textbf{Arch} is an abbreviation for \textit{Architecture}. The \textbf{bold} numbers in the results represent the best scores, whereas the \underline{underlined} numbers indicate the second-best scores.\label{ed_auto}}
\end{table}

\begin{table}[!t]
\centering
\resizebox{\linewidth}{!}{
\begin{tabular}{l|l|l|l|lll|l|l|ll|l}
\toprule[1.3pt]
\textbf{Model}          & \textbf{Scale} & \textbf{Arch} & \textbf{PPL\textdownarrow}    & \textbf{B-1} & \textbf{B-2} & \textbf{B-4} & \textbf{MT} & \textbf{R-L }& \textbf{D-1} & \textbf{D-2} & \textbf{PPR\textdownarrow}     \\ \cmidrule{0-11}
PLATO & -- & DO& --& 39.70 & 31.10& --& --&--&5.30&29.10&--\\
DialogWAE & -- & ED & -- & 32.30 & --&--&--&--&31.30&59.70&--\\ \cmidrule{0-11}
ChatGPT        & 175B  & DO   & 11.41 & \underline{7.58}   & \underline{2.71}   & \underline{0.56}   & \textbf{10.13 } & \underline{8.17}   & 10.98  & 47.20  & \textbf{0.00\%}  \\ \cmidrule{0-11}
ChatGLM        & 6B    & DO   & 17.52 & \textbf{10.54}  & \textbf{3.86 }  & \textbf{0.93}   & 9.14   & \textbf{11.91}  & 9.60   & 42.69  & 12.05\% \\ \cmidrule{0-11}
Flan-T5-XXL    & 13B   & ED   & 16.31 & 3.85   & 1.61   & 0.42   & 6.64   & 5.52   & \underline{14.54}  & 47.59  & \textbf{0.00\%}     \\
FastChat-T5    & 3B    & ED   & \textbf{10.27} & 7.45   & 2.59   & 0.50   & \underline{9.15}   & 7.86   & 9.58   & 41.16  & \underline{0.50\%}   \\ \cmidrule{0-11}
Open-LLaMA     & 7B    & DO   & 21.23 & 6.72   & 2.31   & 0.46   & 5.94   & 5.59   & 11.65  & 38.72  & 64.36\% \\
Vicuna         & 13B   & DO   & 78.66 & 6.13   & 2.11   & 0.42   & 8.89   & 6.96   & 10.15  & 45.18  & 38.55\% \\
Alpaca-Lora    & 7B    & DO   & 28.63 & 6.40   & 2.16   & 0.00   & 6.04   & 5.02   & \textbf{17.49}  & \textbf{61.66}  & 3.41\%  \\
Chinese-Alpaca & 13B   & DO   & 22.23 & 6.52   & 2.18   & 0.38   & 7.49   & 5.93   & 13.06  & 51.02  & 2.01\%  \\
GPT4ALL        & 13B   & DO   & 14.72 & 4.84   & 1.24   & 0.13   & 7.72   & 5.77   & 10.24  & 43.53  & 25.50\% \\ \cmidrule{0-11}
Dolly          & 12B   & DO   & 58.29 & 6.09   & 2.01   & 0.40   & 5.70   & 4.25   & 14.14  & \underline{52.33 } & 74.80\% \\
Oasst-Pythia   & 12B   & DO   & \underline{10.68} & 5.40   & 1.45   & 0.19   & 7.62   & 6.09   & 9.23   & 38.91  & 16.47\% \\ \bottomrule[1.3pt]
\end{tabular}
}
\caption{Automatic evaluation results of LLMs on DailyDialog. \label{dd_auto}}
\end{table}

\begin{table}[!t]
\centering
\resizebox{\linewidth}{!}{
\begin{tabular}{l|l|l|l|lll|l|l|ll|l}
\toprule[1.3pt]
\textbf{Model}          & \textbf{Scale} & \textbf{Arch} & \textbf{PPL\textdownarrow}    & \textbf{B-1} & \textbf{B-2} & \textbf{B-4} & \textbf{MT} & \textbf{R-L }& \textbf{D-1} & \textbf{D-2} & \textbf{PPR\textdownarrow}     \\ \cmidrule{0-11}
PLATO & -- & DO& --& 40.60 & 31.50& --& --&--&2.10&12.10&--\\
CTRLStruct &--&ED &--&31.60&11.90&--&--&16.10&3.20&11.40&--\\ \cmidrule{0-11}
ChatGPT        & 175B  & DO   & 10.97 & \underline{6.36}   & \underline{2.37}   & \textbf{0.52}   & \textbf{9.78}   & \underline{8.42 }  & 9.10   & 40.65  & \textbf{0.00\%}       \\ \cmidrule{0-11}
ChatGLM        & 6B    & DO   & 13.89 & 5.98   & 2.07   & 0.40   & 8.85   & \textbf{8.67}   & 6.85   & 34.86  & 12.05\% \\ \cmidrule{0-11}
Flan-T5-XXL    & 13B   & ED   & 51.50 & \textbf{6.51}   & \textbf{2.53}   & \underline{0.43}   & 6.15   & 7.46   & \underline{12.23}  & 39.82  & \textbf{0.00\%}     \\
FastChat-T5    & 3B    & ED   & \underline{10.61} & 5.53   & 2.00   & \underline{0.43}   & \underline{8.98}   & 7.94   & 7.30   & 33.66  & \underline{0.50\%}  \\ \cmidrule{0-11}
Open-LLaMA     & 7B    & DO   & 15.69 & 4.43   & 1.16   & 0.00   & 5.86   & 5.43   & 7.83   & 28.90  & 64.36\% \\
Vicuna         & 13B   & DO   & 12.53 & 3.20   & 1.01   & 0.14   & 7.30   & 4.82   & 5.88   & 30.12  & 38.55\% \\
Alpaca-Lora    & 7B    & DO   & 17.20 & 4.19   & 1.21   & 0.24   & 6.29   & 4.40   & \textbf{12.28}  & \textbf{50.33}  & 3.41\%  \\
Chinese-Alpaca & 13B   & DO   & 14.95 & 4.93   & 1.66   & 0.29   & 7.70   & 6.21   & 10.18  & \underline{44.62}  & 2.01\%  \\
GPT4ALL        & 13B   & DO   & 11.68 & 2.74   & 0.55   & 0.07   & 6.52   & 4.39   & 7.56   & 35.23  & 25.50\% \\ \cmidrule{0-11}
Dolly          & 12B   & DO   & 29.76 & 4.51   & 1.39   & 0.24   & 5.02   & 4.59   & 10.55  & 41.62  & 74.80\% \\
Oasst-Pythia   & 12B   & DO   & \textbf{9.57}  & 3.34   & 0.69   & 0.07   & 6.58   & 4.66   & 6.48   & 28.56  & 16.47\% \\ \bottomrule[1.3pt]
\end{tabular}
}
\caption{Automatic evaluation results of LLMs on PersonaChat. \label{pc_auto}}
\end{table}

\begin{table}[!t]
\centering
\resizebox{0.75\linewidth}{!}{
\begin{tabular}{l|l|lll|ll}
\toprule[1.3pt]
\textbf{Models} & \textbf{Scale} & \textbf{BLEU} & \textbf{BLEU-1} & \textbf{BLEU-4} & \textbf{Dist-1} & \textbf{Dist-2} \\ \cmidrule{0-6}
  CDialGPT              &   104M     &       --        &   --            &    \textbf{3.20}             &   0.83              &            12.71               \\
   GPT-Novel            &     104M           &    --           &    --             &    \underline{2.71}             &     0.80            &      11.72       \\ \cmidrule{0-6}
ChatGPT         & 175B           & 2.55          & 5.45            & 0.96            & \underline{4.83}            & \textbf{28.84}           \\ \cmidrule{0-6}
ChatGLM         & 6B             & 0.83          & 1.51            & 0.40            & 2.08            & 5.74            \\ \cmidrule{0-6}
Vicuna          & 13B            & 3.84          & 7.84            & 1.58            & 4.70            & \underline{26.72}           \\
Alpaca          & 7B             & 4.79          & 8.75            & 2.33            & \textbf{6.15}            & 25.26           \\
Chinese-Alpaca  & 13B            & 2.88          & 5.78            & 0.35            & 4.09            & 25.84           \\
GPT4ALL         & 13B            & 3.78          & 8.37            & 1.33            & 2.25            & 7.83            \\ \cmidrule{0-6}
Dolly           & 12B            & \textbf{5.30}         & \underline{10.70}           & 2.21            & 4.53            & 20.12           \\
Oasst-Pythia    & 12B            & \underline{5.16}          & \textbf{11.36}           & 1.86            & 2.04            & 7.49  \\ \bottomrule[1.3pt]  
\end{tabular}
}
\caption{Automatic evaluation results of LLMs on LCCC.\label{lccc}}
\end{table}

\subsection{T5-Based models}
T5 \cite{DBLP:journals/jmlr/RaffelSRLNMZLL20}, which stands for Text-To-Text Transfer Transformer, is a transformer-based language model developed by Google Research. Instead of training separate models for different tasks, T5 is trained in a text-to-text pattern. This means that it is trained to perform a wide range of NLP tasks by transforming the input text into a standardized format that specifies the task to be performed. In our evaluation, we select two new fine-tuned versions of T5, namely: Flan-T5-XXL\footnote{https://huggingface.co/google/flan-t5-xxl} and FastChat-T5\footnote{https://huggingface.co/lmsys/fastchat-t5-3b-v1.0}. 
\paragraph{Flan-T5-XXL} Flan-T5 \cite{DBLP:journals/corr/abs-2210-11416} is a fine-tuned version model class of T5 that has been trained on a variety of datasets phrased as instructions. It has shown impressive performance on several benchmarks, demonstrating strong zero-shot, few-shot, and Chain-of-Thought (CoT) \cite{DBLP:conf/nips/Wei0SBIXCLZ22} abilities. Flan-T5-XXL is the largest released checkpoint of this model, boasting a parameter volume of 13B. It inherits the extensive knowledge base of T5 while also being capable of understanding natural language instructions and performing the corresponding tasks.
\paragraph{FastChat-T5} FastChat \cite{DBLP:journals/corr/abs-2306-05685} is an open platform for training, serving, and evaluating large language model based chatbots. And FastChat-T5 is an open-source chatbot trained on this platform by fine-tuning Flan-T5-XL (3B parameters) on user-shared conversations collected from ShareGPT.

\subsection{LLaMA-Based Models}
LLaMA \cite{DBLP:journals/corr/abs-2302-13971} is a collection of foundation language models ranging from 7B to 65B parameters proposed by Meta AI. Unlike other famous LLMs, LLaMA is only trained on publicly avaiable data, making it compatible with open-sourcing. Numerous remarkable and impressive models have emerged as a result, built upon the LLaMA framework and trained using diverse datasets. Among these models, we have chosen a few prominent ones for evaluation: Open-LLaMA, Vicuna, Alpaca, and GPT4ALL.
\paragraph{Open-LLaMA} Open-LLaMA \cite{openlm2023openllama} is an open reproduction of LLaMA trained on the RedPajama dataset \cite{together2023redpajama}. We leverage the 7B version\footnote{https://github.com/openlm-research/open\_llama} of this model  for evaluation.

\paragraph{Alpaca} \cite{alpaca} is fine-tuned based on a 7B LLaMA model using a dataset consisting of 52,000 instances of instruction-following data. This dataset is generated using the techniques outlined in the Self-Instruct paper \cite{DBLP:journals/corr/abs-2212-10560}, which aims to address the limited instruction-following capabilities of LLaMA models. To create the training data, the authors initially generate the data using OpenAI's GPT-3 and subsequently convert it into 52,000 instances of instruction-following conversational data using the Self-Instruct pipeline. This dataset is referred to as the Alpaca dataset. The Alpaca model is then fine-tuned to generate responses in conversations similar to ChatGPT.

In our evaluation, we utilize Alpaca-Lora-7B\footnote{https://huggingface.co/chainyo/alpaca-lora-7b}, a low-rank adapter for LLaMA-7b fit on the Stanford Alpaca dataset, and Chinese-Alpaca-13b\footnote{https://huggingface.co/shibing624/chinese-alpaca-plus-13b-hf}, a Chinese model version of Alpaca.

\paragraph{Vicuna} \cite{zheng2023judging} is fine-tuned based on LLaMA models using user-shared conversations collected from ShareGPT. It is an auto-regressive language model, based on the transformer architecture. So it is basically fine-tuned with ChatGPT conversations. We utilize the 13B version of Vicuna, which is Vicuna-13B\footnote{https://huggingface.co/eachadea/vicuna-13b-1.1}.

\paragraph{GPT4ALL} \cite{gpt4all} is a fine-tuned LLaMA 13B model and the GPT4All community\footnote{https://home.nomic.ai/} has built the GPT4All Open Source datalake as a staging ground for contributing instruction and assistant tuning data for future GPT4All model trains.

\subsection{Pythia-Based Models}
Pythia \cite{DBLP:journals/corr/abs-2304-01373} is a project by EleutherAI\footnote{https://github.com/EleutherAI/pythia} that combines interpret-ability analysis and scaling laws to understand how knowledge develops and evolves during training in autoregressive Transformers. We utilize two versions of Pythia which are Oasst-Pythia and Dolly.

\paragraph{Oasst-Pythia\footnote{https://huggingface.co/OpenAssistant/pythia-12b-sft-v8-7k-steps}} is an open assistant model developed by the Open-Assistant project. It is based on a Pythia 12B model that was fine-tuned on human demonstrations of assistant conversations collected through the Open-Assistant human feedback web app.

\paragraph{Dolly\footnote{https://huggingface.co/databricks/dolly-v2-12b}} is a Language Model (LLM) with 12B parameters, designed to follow instructions accurately. It has been trained on approximately 15,000 instruction/response fine-tuning records known as databricks-dolly-15k. These records were created by Databricks employees and cover various capability domains sourced from InstructGPT \cite{DBLP:conf/nips/Ouyang0JAWMZASR22}. These domains include brainstorming, classification, closed QA, generation, information extraction, open QA, and summarization.

\begin{table}[!t]
\centering
\resizebox{\linewidth}{!}{
\begin{tabular}{l|l|l|l|llll|l|l|l}
\toprule[1.3pt]
\textbf{Model} & \textbf{Scale} & \textbf{Arch} & \textbf{PPL\textdownarrow} & \textbf{B-1} & \textbf{B-2} & \textbf{B-3} & \textbf{B-4} & \textbf{MT} & \textbf{R-L} & \textbf{PPR\textdownarrow} \\ \cmidrule{0-10}
ChatGPT        & 175B  & DO   & \underline{10.86}        & 2.99            & 0.58            & 0.00            & 0.00            & 4.89            & 5.02            & \textbf{0.00\%}          \\
ChatGLM        & 6B    & DO   & 18.56        & 2.80            & 0.87            & 0.25            & 0.00            & 4.80            & 4.91            & 10.78\%      \\ \cmidrule{0-10}
Flan-T5-XXL    & 13B   & ED   & 15.96        & \textbf{5.49}            & \underline{1.21}           & 0.00            & 0.00            & 3.69            & 5.16            & \textbf{0.00\%}          \\ 
FastChat-T5    & 3B    & ED   & \textbf{10.26}        & 2.62            & 0.89            & 0.46            & 0.29            & 4.80            & 4.58            & \underline{0.03\%}       \\ \cmidrule{0-10}
Open-LLaMA     & 7B    & DO   & 45.72        & 0.02            & 0.01            & 0.00            & 0.00            & 0.35            & 0.18            & 73.67\%      \\
Vicuna         & 13B   & DO   & 10.94        & 2.45            & 0.81            & 0.41            & 0.23            & 4.75            & 4.40            & 31.29\%      \\
Alpaca-lora    & 7B    & DO   & 19.22        & 3.41            & 0.56            & 0.00            & 0.00            & 4.19            & 4.23            & 0.13\%       \\
Chinese-Alpaca & 13B   & DO   & 14.30        & \underline{4.40}            & \textbf{1.88}            & \textbf{1.13}            & \textbf{0.74}            & 3.55            & \textbf{10.27}           & 0.15\%       \\
GPT4ALL        & 13B   & DO   & 23.28        & 3.03            & 0.85            & 0.49            & 0.35            & 5.14            & 5.06            & 3.37\%       \\ \cmidrule{0-10}
Dolly          & 12B   & DO   & 15.01        & 3.35            & 1.12            & \underline{0.62}            & 0.40            & \textbf{5.40}            & 6.01            & 46.67\%      \\
Oasst-Pythia   & 12B   & DO   & 18.83        & 3.48            & 1.15            & 0.61            & \underline{0.41}            & \underline{5.23}            & \underline{6.31}            & 0.08\%  \\ \bottomrule[1.3pt]    
\end{tabular}
}
\caption{Automatic evaluation results of LLMs on CNN/DailyMail. \label{cnn_auto}}
\end{table}

\begin{table}[!t]
\centering
\resizebox{\linewidth}{!}{
\begin{tabular}{l|l|l|l|llll|l|l|l}
\toprule[1.3pt]
\textbf{Model} & \textbf{Scale} & \textbf{Arch} & \textbf{PPL\textdownarrow} & \textbf{B-1} & \textbf{B-2} & \textbf{B-3} & \textbf{B-4} & \textbf{MT} & \textbf{R-L} & \textbf{PPR\textdownarrow} \\ \cmidrule{0-10}
ChatGPT        & 175B  & DO   & 14.92        & 7.55            & 2.93            & 1.27            & 0.55            & 11.47           & 10.31           & \textbf{0.00\% }         \\ \cmidrule{0-10}
ChatGLM        & 6B    & DO   & 22.84        & 5.45            & 2.46            & 1.19            & 0.60            & 10.76           & 9.25            & 8.79\%       \\ \cmidrule{0-10}
Flan-T5-XXL    & 13B   & ED   & \textbf{10.90}        & \textbf{12.48}           & \textbf{4.66}            & \textbf{2.19}            & \textbf{1.81}            & \textbf{17.60}          & \textbf{15.06}           & \textbf{0.00\%}        \\
FastChat-T5    & 3B    & ED   & \underline{14.08}        & 8.05            & \underline{3.78}            & 1.83            & 0.78            & \underline{13.22}           & \underline{11.01}           & \textbf{0.00\%}          \\ \cmidrule{0-10}
Open-LLaMA     & 7B    & DO   & 31.13        & 4.57            & 1.31            & 0.55            & 0.00            & 2.31            & 2.70            & 56.79\%      \\
Vicuna         & 13B   & DO   & 14.58        & 7.13            & 3.06            & 1.41            & 0.67            & 12.61           & 10.16           & 30.11\%      \\
Alpaca-lora    & 7B    & DO   & 23.49        & \underline{8.65}            & 2.95            & 1.20            & 0.49            & 10.94           & 9.54            & \underline{1.17\%}       \\
Chinese-Alpaca & 13B   & DO   & 19.21        & 6.65            & 3.31            & \underline{1.88}            & \underline{1.19}            & 5.98            & 8.34            & 5.90\%       \\
GPT4ALL        & 13B   & DO   & 18.79        & 8.47            & 3.46            & 1.68            & 0.95            & 11.73           & 9.81            & 15.79\%      \\ \cmidrule{0-10}
Dolly          & 12B   & DO   & 20.89        & 6.44            & 2.64            & 1.01            & 0.00            & 11.21           & 9.95            & 82.23\%      \\
Oasst-Pythia   & 12B   & DO   & 21.49        & 6.27            & 2.46            & 0.99            & 0.37            & 9.98            & 9.32            & 28.42\%     \\ \bottomrule[1.3pt]
\end{tabular}
}
\caption{Automatic evaluation results of LLMs on XSum.\label{xsum_auto}}
\end{table}


\begin{table}[!t]
\centering
\resizebox{\linewidth}{!}{
\begin{tabular}{l|lll|lllll}
\toprule[1.3pt]
\multirow{2}{*}{\textbf{Models}} & \multicolumn{3}{c|}{\textbf{LCSTS}}               & \multicolumn{5}{c}{\textbf{LOT}}                                                  \\  \cmidrule(lr){2-4} \cmidrule(lr){5-9}
                                 & Rouge-1            & Rouge-2            & Rouge-L            & BLEU              & BLEU-1            & BLEU-4           & Dist-1            & Dist-2            \\ \cmidrule{1-9}
ERNIE                            & --             & --             & \textbf{48.46} & --             & --             & --            & --             & --             \\
RNN-Context                      & \textbf{29.90} & \textbf{17.40} & {\ul 27.20}    & --             & --             & --            & --             & --             \\
LongLM                           & --             & --             & --             & --             & --             & 5.97          & --             & --             \\ \cmidrule{1-9}
ChatGPT                          & 17.20          & 4.92           & 12.05          & {\ul 19.21}    & {\ul 33.34}    & {\ul 8.92}    & {\ul 7.56}     & {\ul 40.87}    \\ \cmidrule{1-9}
ChatGLM                          & {\ul 18.04}    & {\ul 5.88}     & 12.83          & 15.20          & 26.59          & 6.99          & 5.40           & 34.00          \\  \cmidrule{1-9}
Vicuna                           & 16.62          & 4.49           & 11.71          & \textbf{19.48} & \textbf{33.81} & \textbf{9.39} & 7.52           & 37.59          \\
Chinese-Vicuna                   & --             & --             & --             & 13.13          & 26.55          & 4.52          & 5.66           & 32.38          \\
Alpaca                           & 11.52          & 3.51           & 8.52           & 0.63           & 1.08           & 0.33          & 3.18           & 8.02           \\
Chinese-Alpaca                   & 11.98          & 2.42           & 9.05           & 11.91          & 23.68          & 3.89          & 4.85           & 30.10          \\
GPT4ALL                          & 4.13           & 0.93           & 3.05           & 0.94           & 1.83           & 0.40          & 3.86           & 10.02          \\ \cmidrule{1-9}
Dolly                            & 10.83          & 3.84           & 7.41           & 10.09          & 17.43          & 5.13          & \textbf{14.42} & \textbf{45.94} \\
Oasst-Pythia                     & 12.95          & 4.24           & 9.19           & 7.43           & 11.61          & 4.45          & 9.11           & 27.90         \\ \bottomrule[1.3pt]
\end{tabular}
}
\caption{Automatic evaluation results of LLMs on LCSTS and LOT.\label{lcsts}}
\end{table}

\subsection{Dataset}
In our evaluation, we aim to showcase the generation capabilities of LLMs in zero-shot scenarios. Therefore, we refrain from providing any additional information to the model for each of the aforementioned datasets. Specifically:
\begin{itemize}
    \item For datasets of Text Summarization task, we input the text, document, or article to allow the model to extract key information and generate concise summaries. 
    \item For datasets of Dialogue Generation task, we input the dialogue history, enabling the model to generate appropriate responses for the final round of the conversation. 
\end{itemize}

\begin{table*}[!t]
\centering
\resizebox{\linewidth}{!}{
\begin{tabular}{l|lllll|lllll}
\toprule[1.3pt]
\multirow{2}{*}{\textbf{Models}} & \multicolumn{5}{c|}{\textbf{ROCStories}}                                           & \multicolumn{5}{c}{\textbf{WritingPrompts}}                                       \\  \cmidrule(lr){2-6} \cmidrule(lr){7-11}
                                 & B             & B-1            & B-4            & D-1            & D-2            & B              & B-1            & B-4           & D-1            & D-2            \\ \cmidrule{1-11}
MVP                              & --             & --              & \textbf{15.76} & 3.02           & {\ul 75.65}    & --              & --              & --              & --              & --              \\
KEPM                             & --             & \textbf{32.60} & --              & --              & \textbf{78.96} & --              & --              & --             & --              & --              \\
TextBox2.0                       & --             & --              & --              & --              & --              & \textbf{33.79} & --              & --             & --              & \textbf{78.76} \\ \cmidrule{1-11}
ChatGPT                          & 5.70          & 13.60          & 1.41           & \textbf{21.98} & 67.24          & {\ul 4.57}     & \textbf{12.86} & {\ul 0.37}    & 3.66           & 27.92          \\
ChatGLM                          & 0.91          & 2.86           & 0.04           & 6.43           & 36.85          & 4.11           & {\ul 11.13}    & \textbf{0.39} & 2.79           & 20.02          \\ \cmidrule{1-11}
Flan-T5                          & \textbf{9.11} & 17.80          & {\ul 3.73}     & 18.15          & 54.32          & 0.00           & 0.00           & 0.00          & \textbf{14.47} & {\ul 50.95}    \\
FastChat-T5                      & 6.63          & 14.80          & 1.97           & 15.23          & 50.32          & 0.26           & 0.70           & 0.03          & 4.88           & 26.26          \\ \cmidrule{1-11}
Open-LLaMA                       & 3.60          & 9.21           & 0.58           & 14.53          & 46.94          & 0.02           & 0.04           & 0.00          & 4.72           & 22.52          \\
LLaMA2-Chat                      & 4.44          & 10.88          & 0.94           & {\ul 21.02}    & 65.48          & 0.00           & 0.00           & 0.00          & {\ul 10.82}    & 40.09          \\
Vicuna                           & {\ul 7.07}    & 15.31          & 2.23           & 20.17          & 64.82          & 1.69           & 4.56           & 0.17          & 4.48           & 28.53          \\
Chinese-Vicuna                   & 8.88          & {\ul 18.02}    & 3.31           & 20.02          & 63.55          & 0.37           & 0.98           & 0.05          & 5.50           & 29.10          \\
Alpaca                           & 4.69          & 11.64          & 0.99           & 18.02          & 62.61          & 0.08           & 0.22           & 0.01          & 6.16           & 36.23          \\
Chinese-Alpaca                   & 3.77          & 9.30           & 0.78           & 17.89          & 58.42          & 0.65           & 1.85           & 0.05          & 4.08           & 26.41          \\
GPT4ALL                          & 4.61          & 10.99          & 1.17           & 18.87          & 61.51          & 0.73           & 2.03           & 0.07          & 5.18           & 30.70          \\ \cmidrule{1-11}
Dolly                            & 2.81          & 7.04           & 0.50           & 11.31          & 52.03          & 0.24           & 0.68           & 0.02          & 7.07           & 42.52          \\
Oasst-Pythia                     & 2.78          & 7.16           & 0.45           & 10.65          & 48.42          & 0.53           & 1.46           & 0.05          & 4.37           & 27.26    \\ \bottomrule[1.3pt]      
\end{tabular}
}
\caption{Automatic evaluation results of LLMs on ROCStories and WritingPrompts.\label{roc}}
\end{table*}

We defer the evaluation of LLMs on Chinese datasets and other NLG tasks such as story generation, along with results of manual and GPT-4 rating, to future research endeavors.
\subsection{Input Template}
Because LLMs that we evaluate possess the ability to comprehend instructions and perform corresponding tasks, so in order to ensure fairness, we develop an input template that is applied to every dataset for each task, serving as the input for every large language model. This template consists of two components: the instruction and the input. Figure \ref{template} illustrates the templates designed for both the Chinese and English datasets, and Table \ref{input} shows the content of \textit{instruction} and \textit{text} for each dataset.

\subsection{Hyperparameters}
Although each LLM may have its own optimal decoding strategy, for the sake of fairness, we have standardized these hyperparameters across all LLMs. We employ the Top-k and Top-p sampling, with $k=40$ and $p=0.75$. Additionally, a temperature value of $0.2$ and a repetition penalty factor of $1.15$ are imposed. Furthermore, we specify a maximum token length of $512$ and a minimum token length of $10$ for the generated content.

\subsection{Post-Processing Strategy}
Through case study, we observe that despite emphasizing the exclusion of any additional output in the input, regrettably, most LLMs still generate redundant information in their output. Therefore, we find it necessary to apply post-processing to the outputs of these models. To ensure fairness, we adopt the same post-processing strategy for all LLMs. Specifically, we utilize the keywords ``\texttt{\#\#\# response:}'' or ``\texttt{\#\#\#} \begin{CJK*}{UTF8}{gbsn}回复：\end{CJK*}'' for segmentation. If the segmented content consists of a single line, we consider it as the final result. If the segmented content spans multiple lines, we use ``\verb|\n|'' as segmentation keywords and select the first sentence with a length not less than 16 as the final result.

\subsection{Baselines}
There have been numerous previous works on datasets we used, and these works have achieved good results. Therefore, despite the fact that most of these works have proposed models much smaller than LLMs and have predominantly utilized supervised fine-tuning methods, we still compare them with LLMs to highlight some characteristics of LLMs. For each dataset, we select several recent works with better performance and report their results.
\begin{itemize}
    \item For EmpatheticDialogues, we utilize \textbf{EP-PG} \cite{DBLP:conf/acl/LiLBRLK22} that first generates event transition plans and then obtains the final response, and  \textbf{MoEL} \cite{DBLP:conf/emnlp/LinMSXF19} that are consist of one emotion tracker and $n$ emotion listeners.
    \item For DailyDialog, we utilize \textbf{PLATO} \cite{DBLP:conf/acl/BaoHWWW20}, a pre-trained dialogue generation model, and  \textbf{DialogWAE} \cite{DBLP:conf/iclr/GuCHK19}, a conditional wasserstein autoencoder (WAE) specially designed for dialogue modeling.
    \item For PersonaChat, we utilize \textbf{PLATO} as mentioned above, and \textbf{CTRLStruct} \cite{DBLP:conf/www/YinLR23} for dialogue structure learning to effectively explore topic-level dialogue clusters.
\end{itemize}

\subsection{Evaluation Metrics}
\paragraph{Automatic Metrics} 
We utilize several common automatic metrics for NLG tasks. \textbf{PPL} is used to assess the difficulty or confusion of a language model in predicting a sequence of words. \textbf{BLEU} (B-1, B-2, B-3, B-4) \cite{DBLP:conf/acl/PapineniRWZ02} is used to assess the quality of machine-generated translations by comparing them to human reference translations. \textbf{Meteor} (MT) \cite{DBLP:conf/acl/BanerjeeL05} considers the accuracy and recall based on the entire corpus, and get the final measure. \textbf{Rouge-L} (R-L) \cite{lin2004rouge} calculates the overlap between the generated output and the reference summaries or translations using various techniques such as N-gram matching. \textbf{DISTINCT} (D-1, D-2) \cite{DBLP:conf/naacl/LiGBGD16} quantifies how many distinct or different N-grams are present in the generated text, providing an indication of the model's ability to produce varied and non-repetitive output.

Besides these widely-used metrics, we also develop a new metric called \textbf{PostProcess Rate} (PPR), which means the proportion of samples that need to be post-processed to the total number of samples.


\section{Results and Analysis}

\begin{table}[!t]
\centering
\resizebox{0.9\linewidth}{!}{
\begin{tabular}{l|l|lll|ll}
\toprule[1.3pt]
\textbf{Dataset}            & \textbf{Models}  & \textbf{BLEU} & \textbf{BLEU-1} & \textbf{BLEU-4} & \textbf{Dist-1} & \textbf{Dist-2} \\ \cmidrule{1-7}
\multirow{7}{*}{ED}         & Previous SOTA    &     --          & 16.74           & 2.39            & 2.19            & 8.25            \\ \cmidrule{2-7}
                            & ChatGLM-6B       & 2.22          & 6.08            & 0.27            & 3.57            & 22.82           \\ 
                            & + LoRA            & 5.98          & 14.04           & 1.46            & \textbf{9.20}            & \textbf{47.32}           \\
                            & + P-Tuning v2    & \textbf{28.50}         & \textbf{45.66}           & \textbf{15.43}           & 2.59            & 6.93            \\ \cmidrule{2-7}
                            & LLaMA2-Chat-7B   & 1.88          & 5.25            & 0.24            & 4.18            & 27.33           \\
                            & + LoRA           & 4.98          & 9.87            & 1.84            & 4.73            & 20.46           \\ \cmidrule{2-7}
                            & LLaMA2-7B + LoRA & 5.59         & 10.51           & 2.37            & 4.32            & 17.46           \\ \bottomrule[1.3pt]
 \multicolumn{1}{c}{} \\
\toprule[1.3pt]
\multirow{4}{*}{LCCC}       & Previous SOTA    & --            & --              & 3.20            & 0.83            & \textbf{12.71}           \\ \cmidrule{2-7}
                            & ChatGLM-6B       & 0.83          & 1.51            & 0.40            & 2.08            & 5.74            \\
                            & + LoRA            & \textbf{11.73 }        & \textbf{22.38}           & \textbf{5.49}            & \textbf{4.53}            & 8.87            \\
                            & + P-Tuning v2    & 10.38         & 19.92           & 4.71            & 0.84            & 6.31            \\\bottomrule[1.3pt]
 \multicolumn{1}{c}{} \\
\toprule[1.3pt]
\multirow{7}{*}{ROCStories} & Previous SOTA    & --            & --              & 15.76           & 3.02            & \textbf{75.65}           \\ \cmidrule{2-7}
                            & ChatGLM-6B       & 0.91          & 2.86            & 0.04            & 6.43            & 36.85           \\
                            & + LoRA            & \textbf{37.07}         & 55.68           & \textbf{23.03}           & 8.64            & 18.86           \\
                            & + P-Tuning v2    & 36.29         & \textbf{55.91}           & 21.56           & 5.63            & 32.70           \\ \cmidrule{2-7}
                            & LLaMA2-Chat-7B   & 4.44          & 10.88           & 0.94            & \textbf{21.02}           & 65.48           \\
                            & + LoRA           & 9.61          & 17.63           & 4.34            & 7.86            & 35.89           \\ \cmidrule{2-7}
                            & LLaMA2-7B + LoRA & 9.79          & 17.60           & 4.63            & 8.07            & 36.42           \\\bottomrule[1.3pt]
 \multicolumn{1}{c}{} \\
\toprule[1.3pt]
\multirow{4}{*}{LOT}        & Previous SOTA    & --            & --              & 5.97            & --              & --              \\ \cmidrule{2-7}
                            & ChatGLM-6B       & 15.20         & 26.59           & 6.99            & 5.40            & 34.00           \\
                            & + LoRA            & \textbf{25.17}         & \textbf{38.98}           & \textbf{14.91}           & 13.33           & \textbf{58.53}           \\
                            & + P-Tuning v2    & 17.35         & 30.02           & 9.26            & \textbf{13.42}           & 57.47      \\\bottomrule[1.3pt]      
\end{tabular}
}
\caption{Results of finetuning ChatGLM, and LLaMA2 by LoRA and P-Tuning V2 on Empathetic Dialogues, LCCC, ROCStories, and LOT.\label{finetune}}
\end{table}

\subsection{Dialogue Generation}
The automatic metrics results of LLMs on the three datasets are shown in Table \ref{ed_auto}, \ref{dd_auto}, \ref{pc_auto} and \ref{lccc}. Although automatic metrics cannot fully reflect the performance of the models, we can still draw the following conclusions from them.

First, apart from ChatGPT that has the largest scale of 175B, the two T5-based models consistently outperform others in terms of the \textbf{PPR} metric. This indicates that the generated content of Flan-T5-XXL and FastChat-T5 largely aligns with the instruction requirements stated in the input template: "\textit{without any additional output.}" Interestingly, both of these models follow an encoder-decoder architecture, while all other models follow a decoder-only architecture. This suggests that encoder-decoder models demonstrate superior understanding of input instructions under the same model scale. We speculate that having an encoder allows the model to comprehend the input content effectively, thereby executing the corresponding task more successfully.

Second, Alpaca-Lora consistently ranks either first or second in the richness of output content. Moreover, the models using the same architecture as Alpaca-Lora also achieve higher scores in terms of D-1 and D-2. This indicates that LLAMA-based models are capable of producing more diverse and less repetitive content.

Last, ChatGPT, the model with the largest parameter scale, performs the best overall on all four datasets, securing the first or second position most frequently. This suggests that increasing the parameter size and training data volume of LLMs is consistently one of the most important methods for improving model performance.

\subsection{Text Summarization}
The automatic metrics results of LLMs on the three datasets are shown in Table \ref{cnn_auto}, \ref{xsum_auto} and \ref{lcsts}. Our observations from the two datasets can be summarized as follows:

The Flan-T5 and FastChat-T5 models employ an encoder-decoder architecture, exhibiting remarkable proficiency in instruction comprehension, as evident by their minimal requirement for post-processing. This finding is corroborated by the analysis of dialogue generation. Moreover, our investigation on the XSum dataset reveals that both models surpass other LLMs, consistently attaining top positions across various metrics such as BLEU and ROUGE scores. These impressive results are likely attributed to the inherent strengths embedded within their model structures.

\subsection{Story Generation}
The automatic metrics results of LLMs on the three datasets are shown in Table \ref{lcsts} and \ref{roc}.
These tables provide further analysis of the performance of current LLMs on NLG tasks: when the required generated text is excessively long, the models struggle to follow instructions effectively. This is evidenced by the WritingPrompts dataset from Table~\ref{roc}, where many models have BLEU scores that are close to or equal to zero.

\subsection{Finetune LLMs}
To illustrate the enhancement in the performance of LLMs using parameter-efficient fine-Tuning methods, we employ LoRA~\cite{DBLP:conf/iclr/HuSWALWWC22} and P-Tuning V2~\cite{DBLP:journals/corr/abs-2110-07602} to fine-tune ChatGLM-6B, LLaMA2-7B, and LLaMA2-7B-Chat. The results on the Empathetic Dialogues, LCCC, ROCStories, LOT, and LCSTS are presented in Table~\ref{finetune} and~\ref{finetune2}. As shown in the tables, the scores of various metrics significantly improve after fine-tuning the models compared to the non-fine-tuned results. Furthermore, the results in N-grams Matching metrics (BLEU and Rouge) far surpass the previous SOTA results. This demonstrates that LoRA and P-Tuning V2 can substantially enhance the fitting capability of LLMs to datasets without incurring excessive computational resources.

\section{Conclusion}
In this paper, we conduct a comprehensive assessment of several existing large-scale language models (LLMs) in the context of natural language generation (NLG) tasks. Our evaluation encompasses English and Chinese datasets to examine the multilingual capabilities of these LLMs. The results and analyses from both automatic and manual evaluations of LLMs reveal notable trends and phenomena.

\begin{table}[!t]
\centering
\resizebox{0.7\linewidth}{!}{
\begin{tabular}{l|lll}
\toprule[1.3pt]
\textbf{Models} & \textbf{ROUGE-1} & \textbf{ROUGE-2} & \textbf{ROUGE-L} \\ \cmidrule{1-4}
ChatGLM-6B      & 18.04            & 5.88             & 12.83            \\
+ LoRA          & 38.84            & 22.26            & 36.81            \\
+ P-Tuning v2   & 39.20            & 23.58            & 36.95       \\ \bottomrule[1.3pt]    
\end{tabular}
}
\caption{Results of finetuning ChatGLM on LCSTS. \label{finetune2}}
\end{table}

\section*{Acknowledgements}
This research is supported by the National Natural Science Foundation of China (No.62106105), the CCF-Baidu Open Fund (No.CCF-Baidu202307), the CCF-Zhipu AI Large Model Fund (No.CCF-Zhipu202315), the Scientific Research Starting Foundation of Nanjing University of Aeronautics and Astronautics (No.YQR21022), and the High Performance Computing Platform of Nanjing University of Aeronautics and Astronautics.

\bibliographystyle{ccl}
\bibliography{reference}

\end{document}